\documentclass[cameraready]{Interspeech}

\usepackage{hyperref}
\usepackage{url}
\usepackage{graphicx}
\usepackage{booktabs}
\usepackage{xcolor} %
\usepackage[whole]{bxcjkjatype} %

\usepackage{multirow}
\usepackage{cite}
\usepackage{breakcites}
\usepackage{amsfonts}

\usepackage{tcolorbox}
\tcbuselibrary{listings, skins}

\newtcolorbox{promptbox}{
  colback=gray!10,   %
  colframe=gray!70,  %
  fonttitle=\bfseries,
  boxrule=1pt,
  arc=2mm,
  left=2mm, right=2mm, top=1mm, bottom=1mm,
  before skip=6pt,   %
  after skip=6pt,    %
  width=\linewidth,  %
  fontupper=\small %
}

\usepackage[table]{xcolor}

\title{Llama-Mimi: Exploring the Limits of Flattened Speech Language Modeling}

\author[affiliation={1,2}]{Issa}{Sugiura}
\author[affiliation={3,2}]{Shuhei}{Kurita}
\author[affiliation={2}]{Yusuke}{Oda}
\author[affiliation={4,2}]{Ryuichiro}{Higashinaka}

\address{
    $^1$ Kyoto University, Japan\\
    $^2$ NII LLMC, Japan \\
    $^3$ National Institute of Informatics, Japan \\
    $^4$ Nagoya University, Japan
}

\email{sugiura.issa.q29@kyoto-u.jp}


\usepackage{comment}

\begin{document}

\maketitle

\begin{abstract}
Speech Language Models (SpeechLMs) model tokenized speech to capture both semantic and acoustic information. When neural audio codecs based on Residual Vector Quantization (RVQ) are used as audio tokenizers, they produce multiple discrete tokens per time step, yielding inherently multi-level representations. To process these multi-level tokens together, prior work typically adopts hierarchical architectures to capture this structure.
In contrast, recent progress in NLP has progressively reduced architectural inductive biases, moving toward simpler and more scalable single-Transformer architectures.
In this work, we propose Llama-Mimi, which flattens multi-level RVQ tokens produced by the Mimi neural audio codec into a single sequence and models them autoregressively with a Transformer decoder.
We show that Llama-Mimi outperforms a CSM-based hierarchical model on most tasks and achieves the best performance on acoustic consistency. Our models, code, and speech samples are publicly available.\footnote{\url{https://speed1313.github.io/llama-mimi}}
\end{abstract}

\section{Introduction}
\label{sec:intro}
Speech Language Models (SpeechLMs) formulate speech processing as a language modeling problem by converting waveforms into sequences of discrete tokens and modeling them autoregressively~\cite{lakhotia-etal-2021-generative,borsos2023audiolm}. This paradigm enables a single model to handle diverse speech tasks, including synthesis, recognition, and dialogue~\cite{lakhotia-etal-2021-generative,borsos2023audiolm,hassid2023textually,zhang-etal-2023-speechgpt,maiti2024voxtlm}.

A central challenge in SpeechLM design is how to jointly model semantic and acoustic information within a unified framework. In practice, speech tokenizers based on Residual Vector Quantization (RVQ) produce multiple discrete tokens per time step, yielding inherently multi-level representations. To model this structure efficiently, most existing approaches adopt hierarchical architectures derived from the RQ-Transformer~\cite{lee2022rq-transformer,zhu2024GPST,yang2024uniaudio,defossez2024moshispeechtextfoundationmodel,csm2025sesame}. These designs separate temporal modeling across frames from depth-wise modeling within each frame, reducing the effective sequence length and improving computational efficiency. Hierarchical architectures have been successfully employed in models such as Moshi~\cite{defossez2024moshispeechtextfoundationmodel,labiausse2025hibiki}.

Despite their effectiveness, hierarchical designs introduce architectural complexity, including multi-stage pipelines, specialized token organization, and coordination between multiple components~\cite{arora2025landscapespokenlanguagemodels,chou2025flowslm}. In contrast, the evolution of NLP has progressively reduced architectural inductive biases, converging toward simpler and more scalable single decoder-only Transformers~\cite{radford2019language,brown2020gpt3,openai2024gpt4ocard,kaplan2020scalinglawsneurallanguage,wortsman2023smallscaleproxieslargescaletransformer}. This contrast raises a natural question: can a unified single-Transformer architecture effectively model both semantic and acoustic tokens in SpeechLMs?

Motivated by this perspective, we propose Llama-Mimi, a flattened SpeechLM that integrates the Mimi neural audio codec with the Llama Transformer decoder. Given an input speech signal, Mimi first encodes it into a sequence of multi-level RVQ tokens. These tokens are then flattened into a single one-dimensional sequence and modeled autoregressively by a single Transformer decoder.
By removing explicit hierarchical structure over quantizer levels, the model directly learns global and acoustic dependencies from the flattened sequence, reducing architectural inductive biases while maintaining scalability.

We evaluate this design by training both flattened and hierarchical models on similar settings for the continual audio generation task. Llama-Mimi outperforms its hierarchical counterpart on most evaluation tasks and achieves the strongest acoustic consistency among the evaluated models. However, comparisons with phonetic-token-based approaches reveal weaker linguistic performance, highlighting a trade-off between acoustic fidelity and linguistic efficiency in flattened designs.
Finally, we conduct extensive ablation studies on semantic token loss weighting, model size, and number of quantizers.
\begin{figure*}[t]
\centering
\includegraphics[width=0.8\linewidth]{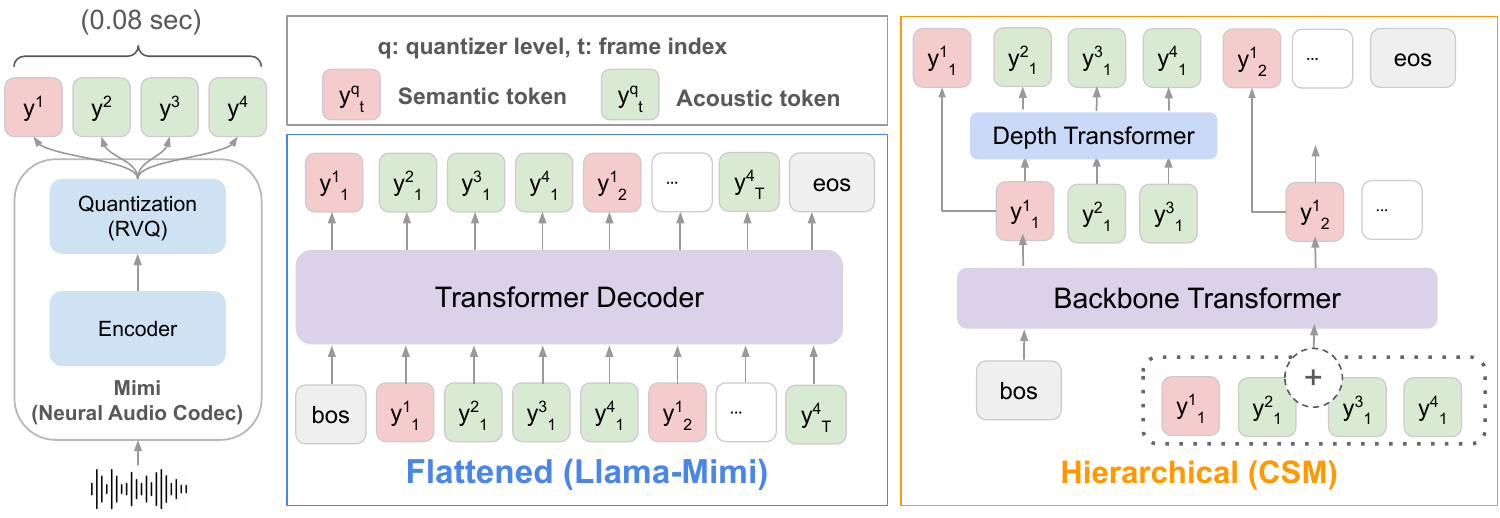}
\caption{Flattened (Llama-Mimi, Ours) vs. hierarchical (CSM~\cite{csm2025sesame}) architectures.
In Llama-Mimi, discrete audio tokens from
Mimi~\cite{defossez2024moshispeechtextfoundationmodel} are flattened
into a single sequence and modeled by a single Transformer decoder,
whereas the hierarchical model separates temporal and depth-wise
modeling across two Transformer decoders.}
\label{fig:llama_mimi}
\end{figure*}

\section{Related Work}

Recent progress in SpeechLM has been driven by the use of discrete audio representations~\cite{lakhotia-etal-2021-generative,kharitonov2022pgslm,borsos2023audiolm,hassid2023textually,rubenstein2023audiopalmlargelanguagemodel,copet2024musicgen}.
Early studies such as GSLM explored phonetic-token–based speech language modeling~\cite{lakhotia-etal-2021-generative,hassid2023textually}. Their approach first converts speech waveforms into sequences of discrete phonetic tokens by quantizing the outputs of self-supervised learning (SSL) based speech encoders~\cite{schneider19wav2vec,wei2021hubert,sanyuan2022wavlm,chung2021w2v-BERT} via k-means clustering. These token sequences are then modeled autoregressively with a Transformer decoder~\cite{vaswani2017transformer}, and the predicted tokens are finally converted back into speech waveforms using a neural vocoder~\cite{jonathan2018tacotron2,jungil2020HiFiGAN}.
While this pipeline enables language-model–style training for speech, tokens derived from SSL encoders primarily capture phonetic content and encode only limited prosodic information~\cite{sicherman2023analysing}. As a result, such models have limited capacity for rich acoustic modeling.

To better preserve acoustic detail, subsequent work adopted neural audio codecs~\cite{neil2021soundstream,defossez2024moshispeechtextfoundationmodel}.
Neural audio codecs compress speech by training a VQ-VAE~\cite{oord2017vqvae} model to reconstruct the waveform, thereby learning quantized latent representations that enable accurate reconstruction with minimal error.
A commonly adopted technique in high-performance neural audio codecs is Residual Vector Quantization (RVQ)~\cite{vasuki2006residualvector}, which performs multi-stage quantization by successively encoding residual errors. This process produces multiple discrete tokens per frame in a coarse-to-fine-grained manner, introducing a new modeling challenge: how to effectively handle such multi-level token structures within an autoregressive decoder.
A common solution is to adopt a hierarchical architecture based on the RQ-Transformer~\cite{lee2022rq-transformer}, which employs dual Transformer decoders along the temporal (frame) and depth (quantizer) dimensions.
This design reduces effective sequence length while modeling multi-level tokens efficiently~\cite{defossez2024moshispeechtextfoundationmodel,csm2025sesame}.
However, such hierarchical designs require multiple model components, increasing architectural complexity, hyperparameter tuning overhead, and potential optimization difficulty.

In this work, inspired by the success of single-Transformer architectures in NLP~\cite{radford2019language,brown2020gpt3}, we instead explore a flattened approach that linearizes multi-level tokens within each frame into a single sequence and models them using a single Transformer decoder.
Although similar flattening strategies have been adopted in prior work
such as SpiritLM~\cite{nguyen2025spirit}, SpiritLM interleaves text
tokens with SSL-based phonetic tokens. To the best of our knowledge, this is the
first study to systematically evaluate a flattened SpeechLM that relies solely on tokens from an RVQ-based neural audio codec.

\section{Llama-Mimi}

\subsection{Background: Mimi}

Mimi~\cite{defossez2024moshispeechtextfoundationmodel} is an RVQ-based neural audio codec~\cite{neil2021soundstream,defossez2023encodec} that follows the unified tokenizer framework of SpeechTokenizer~\cite{zhang2024speechtokenizer}. In this framework, the first level of RVQ is semantically distilled from the self-supervised model WavLM~\cite{sanyuan2022wavlm}, such that the first quantizer primarily captures high-level linguistic information (semantic tokens), while the remaining quantizers encode residual acoustic details (acoustic tokens). In the released model, Mimi converts a 24kHz waveform into RVQ codes at 12.5 frames per second, with each codebook containing 2048 entries and supporting up to 32 quantizers.

\subsection{Llama-Mimi Architecture}
Llama-Mimi is a speech language modeling architecture that integrates the Mimi with a Transformer decoder, as illustrated in Figure~\ref{fig:llama_mimi}.
Given a single-channel waveform $\mathbf{x} \in \mathbb{R}^T$, we encode it into discrete audio tokens using Mimi:
\begin{equation}
\mathbf{h} = \mathrm{Mimi}_{\text{enc}}(\mathbf{x}) = (y^1_1, y^2_1, \dots, y^Q_1, y^1_2, \dots, y^Q_{T'}),
\end{equation}
where $Q$ denotes the number of quantizers, $T'$ ($T' \ll T$) the number of frames, and $y^q_t$ the token produced by the $q$-th quantizer at time $t$.
We then model this audio token sequence autoregressively using a Transformer decoder.
Specifically, we adopt Llama~3~\cite{grattafiori2024llama3herdmodels} as the backbone and train it with the standard next-token prediction objective~\cite{radford2019language}.
To enable Llama~3 to process audio tokens, we extend its vocabulary to include all RVQ tokens as well as two special tokens, \texttt{<audio>} and \texttt{</audio>}.
Within each frame, semantic tokens are predicted before acoustic tokens,
allowing the latter to be conditioned on higher-level linguistic representations.
This ordering follows the coarse-to-fine-grained structure of RVQ.
This ordering encourages the model to generate speech that is both linguistically coherent and acoustically detailed.

At inference, Llama autoregressively generates audio tokens
$\hat{\mathbf{h}}$ conditioned on the given context.
The generated tokens are decoded into an audio waveform $\hat{\mathbf{x}} = \mathrm{Mimi}_{\text{dec}}(\hat{\mathbf{h}})$, and generation stops when the \texttt{</audio>} token is produced.

\begin{table*}[t]
\caption{Performance of each model across tasks. Results marked with $\dag$ are taken from the original papers. Best scores among models with around 1B parameters are \textbf{bolded}. SSL and RVQ refer to self-supervised and residual vector quantization–based audio tokenizers.}
\label{tab:likelihood_based_result}
\setlength{\tabcolsep}{1pt}
\centering
\footnotesize
\begin{tabular}{lrccccccccccccc}
\toprule
\textbf{Model} & \textbf{Train} & \textbf{Tokenizer} & \multicolumn{7}{c}{\textbf{Acoustic}} & \multicolumn{3}{c}{\textbf{Linguistic}} & \multicolumn{2}{c}{\textbf{Generation}}\\
& \textbf{Data}& & \multicolumn{7}{c}{\textbf{SALMon}} & \textbf{sWUGGY} & \textbf{sBLIMP} & \textbf{T-Story}&\textbf{Speaker}&\textbf{Content} \\
& \textbf{Size} & & \multicolumn{5}{c}{\textbf{Consistency}} & \multicolumn{2}{c}{\textbf{Alignment}} & & & \textbf{Cloze} & \textbf{Sim.}&\textbf{Quality}\\
 \cmidrule(lr){4-8} \cmidrule(lr){9-10}
& (Hours) & & \textbf{Sentiment} & \textbf{Speaker} & \textbf{Gender} & \textbf{BG (all)} & \textbf{Room} & \textbf{Sentiment} & \textbf{BG} & & & & \\
\midrule
SpiritLM-base-7B & 570K&SSL & 54.0&68.0&67.0&56.0&54.5 &46.0&50.5&66.9&55.8& 84.6 & 0.035&4.43\\
SpiritLM-exp-7B & 570K & SSL & 72.0 & 80.5 & 85.5 & 64.0 & 55.5 & 52.5 & 59.5 & 64.7 & 54.6& 75.7 & 0.078& 3.69\\
Moshi-7B (Base) & 7,000K & RVQ& -- & -- & -- & -- & -- & -- & -- & 74.3\rlap{$^\dag$} & 58.9\rlap{$^\dag$} & 81.8\rlap{$^\dag$}  & -- & --\\
\midrule
Flow-SLM-1B-ext & 110K & RVQ & 62.5	&76.0	&80.0& 69.0 & 73.0 & \textbf{55.0} & 53.0 & 70.7 & \textbf{60.6} & 65.6 & 0.331 & 3.12\\
TWIST-1.3B &150K &  SSL&  61.5 & 69.0 & 69.5 & 60.5 & 59.0 & 53.0 & \textbf{56.5} & \textbf{71.7} & 56.8 & \textbf{69.9} & 0.110 & \textbf{3.63} \\
\rowcolor{gray!10}
\multicolumn{15}{l}{\textit{Our Controlled Comparison}} \\
CSM-1.3B & 240K & RVQ & 73.5 & 77.5 & 75.0 & 70.5 & 81.5 & 45.5 & 54.0 & 63.7 & 53.4&58.0 & 0.320 & 2.80\\
Llama-Mimi-1.3B & 240K & RVQ& \textbf{79.0} & \textbf{85.0} & \textbf{83.5} & \textbf{73.5} & \textbf{92.0} & 48.5 & 53.5 & 68.7 & 54.3 & 64.0 & \textbf{0.346} & 3.01\\
\bottomrule
\end{tabular}
\end{table*}

\section{Experiments}
Our goal is to explore the potential of flattened speech language modeling.
To this end, we train both a flattened model (Llama-Mimi) and a hierarchical counterpart (CSM)~\cite{csm2025sesame}, as shown in Figure~\ref{fig:llama_mimi}.
Both models are trained with comparable parameter sizes on the same dataset under closely matched training settings, and are further compared with existing baselines.

\subsection{Training Settings}
\noindent\textbf{Models.}
For the flattened approach, we follow the Llama-Mimi architecture and train a model using Llama-3.2-1B~\cite{grattafiori2024llama3herdmodels} as the backbone. We refer to this model as Llama-Mimi-1.3B.
For the hierarchical approach, we use a CSM architecture~\cite{csm2025sesame}, which consists of a Llama-3.2-1B backbone Transformer and a depth Transformer with approximately 100M parameters. We refer to this model as CSM-1.3B.
Both models use the Mimi neural audio codec~\cite{defossez2024moshispeechtextfoundationmodel} as a unified audio tokenizer.
To balance sequence length and reconstruction quality, we use $Q=4$ quantizers, yielding 50 tokens per second.
Following TWIST~\cite{hassid2023textually}, the backbone Transformers are initialized from pretrained checkpoints, while the parameters of Mimi are kept frozen during training.

\noindent\textbf{Datasets.}
We train both models on the same collection of publicly available speech corpora, including Libri-Light~\cite{librilight}, The People's Speech~\cite{galvez2021the}, VoxPopuli~\cite{wang-etal-2021-voxpopuli}, and Emilia~\cite{he2024emilia}.
Restricting to the English subsets, the combined training data amounts to approximately 240k hours of audio.
All audio samples are truncated to a maximum duration of 20 seconds
to limit the sequence length and reduce GPU memory consumption during training.

\noindent\textbf{Training hyperparameters.}
The global batch size is set to 1,024, and the maximum sequence length is fixed at 1,024 tokens.
We train for 100,000 steps with a learning rate that linearly warms up to $3 \times 10^{-4}$ over the first 1,500 steps, remains constant for 80\% of training, and then linearly decays to $3 \times 10^{-5}$ over the final 20\%.
All experiments are conducted on 32 NVIDIA H200 GPUs using FSDP~\cite{zhao2023pytorchfsdpexperiencesscaling}.
Training Llama-Mimi-1.3B takes approximately 48 hours under this setup.

\subsection{Evaluation Settings}

\noindent\textbf{Likelihood-based evaluation.}
For acoustic evaluation, we use SALMon~\cite{maimon2025salmonsuiteacousticlanguage}, which consists of two criteria: acoustic consistency and acoustic–semantic alignment.
Acoustic consistency measures whether the model assigns higher likelihood to natural utterances than to versions with unnatural acoustic perturbations.
Acoustic–semantic alignment evaluates whether the model prefers samples in which acoustic signals match their semantic content over mismatched ones.

For linguistic evaluation, we use sWUGGY~\cite{nguyen2020zeroresourcespeechbenchmark}, sBLIMP~\cite{nguyen2020zeroresourcespeechbenchmark}, and T-StoryCloze~\cite{hassid2023textually}.
sWUGGY tests whether the model assigns higher probability to real words than to pseudo-words, while sBLIMP measures preference for grammatically correct sentences.
T-StoryCloze evaluates the coherence of a continuation given a spoken audio prompt.
Following AudioLM~\cite{borsos2023audiolm}, we compute perplexity only over semantic tokens for these linguistic tasks.

\begin{table*}[t]
\caption{Speech generation examples with prompts and model outputs (audio transcribed using Whisper Turbo).}
\label{tab:continuation}
\footnotesize
\setlength{\tabcolsep}{2pt}
\centering
\begin{tabular}{p{1.5cm}|p{8cm}|p{6.5cm}}
\toprule
\textbf{Prompt} & \textbf{Llama-Mimi-1.3B} & \textbf{Llama-Mimi-8B}\\
\midrule
\textcolor{teal}{Hi. How are you?} &  \textcolor{teal}{Hi. How are you?} I guess this is a thing indifferent to you. When doing the second call, you bring me back to the gate. You know that it wasn't too dark to mine, but when it took you from my side, what?   &  \textcolor{teal}{Hi. How are you?} I'm astonished to see you. Why, I've been looking at you and you had put on yourself out pretty well. And what do you suppose I did there last night?\\
\midrule
\textcolor{teal}{1,2,3,4,5,6} & \textcolor{teal}{1,2,3,4,5,6},7,9 ominous in the stream thing of manual reading and genius and personal discoveries and daily vehicles and data scientific studies and advice and probably hymns [...] & \textcolor{teal}{1, 2, 3, 4, 5, 6}, 7, 8, 9, 10, 11, 12, 9, 8. \\

\bottomrule
\end{tabular}
\end{table*}

\begin{table*}[t]
\caption{Effect of semantic token loss weighting in Llama-Mimi-1.3B.}
\label{tab:semantic_token_weight}

\footnotesize
\setlength{\tabcolsep}{3pt}
\centering
\begin{tabular}{ccccccccccccc}
\toprule
 \textbf{Semantic} & \multicolumn{7}{c}{\textbf{SALMon}} & \textbf{sWUGGY} & \textbf{sBLIMP} & \textbf{T-Story} & \textbf{Speaker} & \textbf{Content}\\
\textbf{Token Weight}& \multicolumn{5}{c}{\textbf{Consistency}} & \multicolumn{2}{c}{\textbf{Alignment}} & & &\textbf{Cloze} &\textbf{Sim.}&\textbf{Quality} \\
\cmidrule(lr){2-6} \cmidrule(lr){7-8}
\textbf{$\lambda$}&  \textbf{Sentiment} & \textbf{Speaker} & \textbf{Gender} & \textbf{BG (all)} & \textbf{Room} & \textbf{Sentiment} & \textbf{BG} & &&& &\\
\midrule
  1& \textbf{79.0} & \textbf{85.0} & \textbf{83.5} & \textbf{73.5} & \textbf{92.0} & 48.5 & \textbf{53.5} & 68.7 & 54.3 & 64.0 & \textbf{0.346} & 3.01\\
 100 & 65.0 & 71.5 & 76.0 & 65.0 & 74.0 & \textbf{53.5} & 52.5 & \textbf{68.8}& \textbf{55.4}& \textbf{68.4} &  0.196 & \textbf{3.87}\\
\bottomrule
\end{tabular}
\end{table*}

\begin{table}[t]
\caption{Effect of model size on Llama-Mimi.}
\label{tab:model_scale}
\footnotesize
\setlength{\tabcolsep}{3.5pt}
\centering
\begin{tabular}{cccccc}
\toprule
\textbf{Model Size} & \textbf{sWUGGY} & \textbf{sBLIMP} & \textbf{T-Story} & \textbf{Speaker} & \textbf{Content} \\
& & &\textbf{Cloze} & \textbf{Sim.} & \textbf{Quality}\\
\midrule
1.3B&  68.7 & 54.3 & 64.0&0.346 & 3.01\\
8B & \textbf{68.8} & \textbf{55.1} & \textbf{67.6} & \textbf{0.348} & \textbf{4.03} \\
\bottomrule
\end{tabular}
\end{table}

\begin{table}[t]
\caption{Effect of the number of quantizers (Llama-Mimi-1.3B).}
\label{tab:ablation_quantizer_num}
\footnotesize
\setlength{\tabcolsep}{4pt}
\centering
\begin{tabular}{ccccccc}
\toprule
\textbf{\#Quantizers} &  \multicolumn{4}{c}{\textbf{Audiobox-Aesthetics}}& \textbf{Speaker} &  \textbf{Content}\\
$Q$& \textbf{PQ} & \textbf{PC} & \textbf{CE} & \textbf{CU} & \textbf{Sim.}& \textbf{Quality}\\
\midrule
 2 & 5.02& \textbf{1.62} &4.58 & 4.72& 0.201 & \textbf{3.53}  \\
 4 &5.55 & \textbf{1.62} & 5.09& 5.26 & 0.346 & 3.01\\
 8 & \textbf{6.01}& 1.61& \textbf{5.40} & \textbf{5.66}& \textbf{0.474} & 2.54 \\
\bottomrule
\end{tabular}
\end{table}

\noindent\textbf{Generation-based evaluation.}
We evaluate speech generation in a continuation setting: given a short audio prompt, the model generates subsequent speech.
The outputs are assessed in terms of speaker consistency and spoken content quality.
We use the LibriSpeech test-clean set~\cite{librilight} to prepare prompts and reference continuations. For each audio sample, the first three seconds serve as the prompt, and the model generates up to 20 seconds of continuation.
Sampling is performed with temperature 0.8 and top-$k$ 30.
Speaker consistency is evaluated using the pyannote speaker embedding model~\cite{Bredin2020pyannotate}, measured as the cosine similarity between the prompt and continuation.

For spoken content quality, we adopt the LLM-as-Judge framework~\cite{zheng2023judgingllmasajudgemtbenchchatbot}; We transcribe the prompt and continuation using Whisper Turbo~\cite{radford2023whisper}, and evaluate the continuation with GPT-4o (\texttt{gpt-4o-2024-11-20})~\cite{openai2024gpt4ocard} on a 1–10 scale based on relevance, coherence, fluency, and informativeness.

\noindent\textbf{Baselines.}
We compare our models with recent SpeechLMs: TWIST-1.3B~\cite{hassid2023textually}, Flow-SLM-1B-ext~\cite{chou2025flowslm}, Moshi-7B~\cite{defossez2024moshispeechtextfoundationmodel}, and SpiritLM-7B~\cite{nguyen2025spirit}.
TWIST-1.3B converts speech into phonetic token sequences using HuBERT, which are then modeled by a language model.
Flow-SLM-1B-ext predicts the next pre-quantization embedding in Mimi using Conditional Flow Matching (CFM)~\cite{lipman2023flow}.
SpiritLM-base-7B models an interleaved token stream of text tokens and HuBERT-based phonetic tokens. We also evaluate SpiritLM-exp-7B, which incorporates additional pitch and style tokens.
Moshi-7B employs the RQ-Transformer architecture for multi-stream modeling, jointly processing the user and system streams, and further incorporates the ``Inner Monologue'' strategy, in which audio tokens are conditioned on intermediate text tokens to improve linguistic performance.
Since the base model of Moshi-7B is not publicly available, we report the results from the original paper.

\subsection{Main Results}
Tables~\ref{tab:likelihood_based_result} present the performance of each model across tasks.
Llama-Mimi-1.3B outperforms CSM-1.3B on all tasks except for the background alignment task in SALMon.
This indicates that the flattened modeling approach yields stronger overall performance than the hierarchical baseline.
We attribute this advantage to the use of a single Transformer decoder
in Llama-Mimi-1.3B, which allows all semantic and acoustic tokens to attend to each other directly, enabling the model to capture fine-grained cross-level dependencies that are constrained by the two-stage processing in hierarchical designs.

Compared with the baseline models, Llama-Mimi-1.3B achieves the best scores on the acoustic consistency tasks and also performs strongly on speaker similarity, indicating its strong ability to capture acoustic details.
However, on linguistic tasks and the spoken content quality task, Llama-Mimi-1.3B underperforms TWIST-1.3B and Flow-SLM-1B-ext. We hypothesize that this gap primarily arises from the longer sequence length induced by flattening RVQ tokens. Because Llama-Mimi models both semantic and acoustic tokens autoregressively in a single sequence, it must process more tokens than approaches that focus primarily on semantic content.
Specifically, TWIST-1.3B relies on SSL-based phonetic tokens, which are shorter and more semantically focused, enabling more efficient semantic modeling. Similarly, Flow-SLM-1B-ext predicts one frame-level context vector per step using a Transformer to produce a pre-quantization embedding per frame and thereby avoiding the $Q$-fold increase in sequence length that occurs with the token-level flattening in Llama-Mimi.

\subsection{Ablation Studies}
To better understand the limitations identified in the main results, we conduct ablation studies on Llama-Mimi, analyzing which design choices contribute to mitigating the bottlenecks in linguistic performance and spoken content quality.

\noindent\textbf{Effect of semantic token loss weighting.}
Our main results indicate that Llama-Mimi's performance on linguistic tasks remains relatively weak. Prior work~\cite{defossez2024moshispeechtextfoundationmodel} reports that assigning a larger loss weight to semantic tokens than to acoustic tokens improves linguistic task performance. Motivated by this finding, we adopt the same strategy and analyze the effect of semantic token loss weighting.
Specifically, we optimize the following weighted objective:
\begin{equation}
L = \lambda L_{\text{semantic}} + L_{\text{acoustic}},
\end{equation}
where $\lambda$ controls the relative weight of semantic tokens.
We compare two settings with $\lambda = 1$ and $\lambda = 100$.

The results are shown in Table~\ref{tab:semantic_token_weight}. Increasing the semantic token weight from 1 to 100 improves performance on linguistic tasks, while degrading metrics related to acoustic consistency and speaker similarity. This behavior reveals a trade-off between semantic accuracy and acoustic quality.

\noindent\textbf{Effect of model size.}
To study the effect of model size on Llama-Mimi, we additionally train Llama-Mimi-8B, using Llama-3.1-8B as the backbone, while keeping all other training settings identical to Llama-Mimi-1.3B.
The results are reported in Table~\ref{tab:model_scale}.
Scaling the model size from 1.3B to 8B consistently improves performance across all tasks.
In particular, substantial improvements are observed in the quality of spoken content, indicating that Llama-Mimi-8B is more effective at generating natural and coherent speech continuations.

For qualitative analysis, Table~\ref{tab:continuation} presents continuations generated by Llama-Mimi-1.3B and 8B using audio prompts from the TWIST demo~\cite{hassid2023textually}.
The 8B model generates continuations that more closely follow the semantic content and intent of the prompts, whereas the 1.3B model produces less consistent continuations.
These observations suggest that larger models are more effective at handling long sequences.

\noindent\textbf{Effect of the number of quantizers.}
The number of quantizers determines both the effective sequence length and the level of acoustic detail encoded in the representation.
To analyze this effect, we vary the number of quantizers ($Q \in \{2,4,8\}$) and evaluate generation quality on LibriSpeech test-clean. In addition to speaker similarity and spoken content quality, we assess audio quality using the Audiobox-Aesthetics model~\cite{tjandra2025metaaudioboxaestheticsunified}, which scores audio along four dimensions.

Table~\ref{tab:ablation_quantizer_num} shows that increasing the number of quantizers consistently improves audio quality and speaker similarity, but this improvement comes at the cost of degraded spoken content quality. Notably, when $Q=2$, the model achieves spoken content quality comparable to TWIST-1.3B, indicating that a smaller number of quantizers better preserves linguistic information. A possible explanation is that increasing the number of quantizers lengthens the token sequence and shifts modeling capacity toward low-level acoustic reconstruction, making it more difficult to accurately capture higher-level linguistic content.

\section{Conclusion}
We introduced Llama-Mimi, which models multi-level discrete tokens produced by Mimi within a unified single-sequence framework using a single Transformer decoder. We showed that Llama-Mimi outperforms its CSM-based hierarchical
counterpart on most evaluation tasks and achieves the best acoustic
consistency among evaluated models. At the same time, it exhibits lower performance on linguistic tasks compared to SSL-driven phonetic-token-based approaches, revealing a trade-off between acoustic fidelity and linguistic efficiency.
We hope these findings provide insights into the design of SpeechLMs.

\vfill\pagebreak

\section{Acknowledgments}
We used ABCI 3.0 provided by AIST and AIST Solutions with support from ``ABCI 3.0 Development Acceleration Use''.

\section{Generative AI Use Disclosure}
Generative AI was employed solely for the purpose of checking typographical errors and refining the English expression in this manuscript. The final content was thoroughly reviewed and verified by the authors.

\bibliographystyle{IEEEtran}
\bibliography{mybib}

\end{document}